\documentclass{article} % For LaTeX2e
\usepackage{iclr2020_conference,times}

\usepackage{amsmath,amsfonts,bm}

\def\eqref#1{equation~\ref{#1}}
\def\1{\bm{1}}

\DeclareMathAlphabet{\mathsfit}{\encodingdefault}{\sfdefault}{m}{sl}
\SetMathAlphabet{\mathsfit}{bold}{\encodingdefault}{\sfdefault}{bx}{n}

\usepackage{hyperref}
\usepackage{url}

\usepackage{amsthm}
\usepackage{graphicx}  %Required
\usepackage{booktabs} % For formal tables
\usepackage{multirow}
\usepackage{mathrsfs}
\usepackage{amsmath}
\usepackage{amsfonts}
\usepackage{balance}
\usepackage{enumitem}
\usepackage{xcolor}
\usepackage{balance}
\usepackage{wrapfig}
\usepackage{subcaption}
\usepackage{enumitem}
\usepackage{lipsum}
\usepackage{wrapfig}
\usepackage{xcolor}
\usepackage{amssymb}
\usepackage{wasysym}
\usepackage{siunitx}
\usepackage[normalem]{ulem}
\usepackage{xcolor}
\makeatletter
\def\squiggly{\bgroup \markoverwith{\textcolor{red}{\lower3.5\p@\hbox{\sixly \char58}}}\ULon}
\makeatother

\title{Attention Interpretability Across NLP Tasks}

\author{Shikhar Vashishth\thanks{Work done during an internship at Google} \\
Indian Institute of Science\\
\small \texttt{shikhar@iisc.ac.in}
\And
Shyam Upadhyay \\ 
Google Assistant \\
\small \texttt{shyamupa@google.com}\\
\And
Gaurav Singh Tomar \\
Google Research \\
\small \texttt{gtomar@google.com}
\And Manaal Faruqui \\
Google Assistant\\
\small \texttt{mfaruqui@google.com}
}

 \iclrfinalcopy % Uncomment for camera-ready version, but NOT for submission.
\begin{document}

\maketitle

%\gst{this is a comment from gaurav}
%\mf{this is a comment from manaal}
%\su{this is a comment from shyam}
%\sv{this is a comment from shikhar}

\newcommand{\mo}[1]{\mathcal{#1}}
\newcommand{\bmm}[1]{\bm{\mathcal{#1}}}
\newcommand{\real}[1]{\mathbb{R}^{#1}}

\newtheorem{theorem}{Theorem}[section]
\newtheorem{claim}[theorem]{Claim}
\newtheorem{lemma}[theorem]{Lemma}
\newtheorem{observation}[theorem]{Observation}
\newtheorem{proposition}[theorem]{Proposition}
\begin{abstract}
%Attention mechanism has become an integral part of all deep learning architectures for its performance benefit. 
The attention layer in a neural network model provides insights into the model's reasoning behind its prediction, which are usually criticized for being opaque. Recently, seemingly contradictory viewpoints have emerged %\su{cite something here} \sv{Not sure whether people cite papers in abstract} 
about the interpretability of attention weights \citep{jain-wallace-2019-attention,vig19}. Amid such confusion arises the need to understand attention mechanism more systematically. In this work, we attempt to fill this gap by giving a comprehensive explanation which justifies both kinds of observations (i.e., when is attention interpretable and when it is not). 
%: We find that attention weights are interpretable and correlate with feature importance measures when weights are not computed in isolation based on the input and are crucial for model output. 
% for tasks where attention plays a substantial role in model's prediction. 
Through a series of experiments on diverse NLP tasks, we validate our observations and reinforce our claim of interpretability of attention through manual evaluation. 
\end{abstract}

\section{Introduction}
\label{sec:introduction}

%\begin{wrapfigure}{R}{0.3\textwidth}
%	\begin{center}
%		\includegraphics[width=0.3\textwidth]{images/motivation_plot-crop.pdf}
%	\end{center}
%	\caption{\label{fig:motivation_plot}Comparison of performance with and without neural attention on text classification (IMDB) and Natural Language Inference tasks (SNLI). The results show that attention does not substantially effect performance on text classification. However, the same does not hold for NLI. Please refer to Section \ref{sec:introduction} for more details. 
%	%\sv{This figure can be removed.} \su{we can keep this figure, but we should add NMT also.}
%	}
%\end{wrapfigure}
Attention is a way of obtaining a weighted sum of the vector representations of a layer in a neural network model \citep{bahdanau_iclr14}.
%Attention mechanism  since its inception has become an integral part of deep learning architectures. 
It is used in diverse tasks ranging from machine translation \citep{luong_att}, language modeling \citep{app_lang_model} to image captioning \citep{app_img_caption}, and object recognition \citep{app_obj_recog}. Apart from substantial performance benefit \citep{vaswani_attention_is_all}, attention also provides interpretability to neural models \citep{atten_interpret1,atten_interpret2,atten_interpret3} which are usually criticized for being black-box function approximators \citep{black_box_nn}.

There has been substantial work on understanding attention in neural network models. On the one hand, there is work on showing that attention weights are not interpretable, and altering them does not significantly affect the prediction \citep{jain-wallace-2019-attention,smith_acl19}. While on the other hand, some studies have discovered how attention in neural models captures several linguistic notions of syntax and coreference \citep{vig19,bert_attention_manning19,dipanjan_bert_acl}. Amid such contrasting views arises a need to understand the attention mechanism more systematically. In this paper, we attempt to fill this gap by giving a comprehensive explanation which justifies both kinds of observations. %\mfar{Not all of these are neural language models. Did you mean neural NLP models? Or simply neural network models?} \sv{Corrected}

\begin{figure*}[t!]
	\centering
	\includegraphics[width=\linewidth]{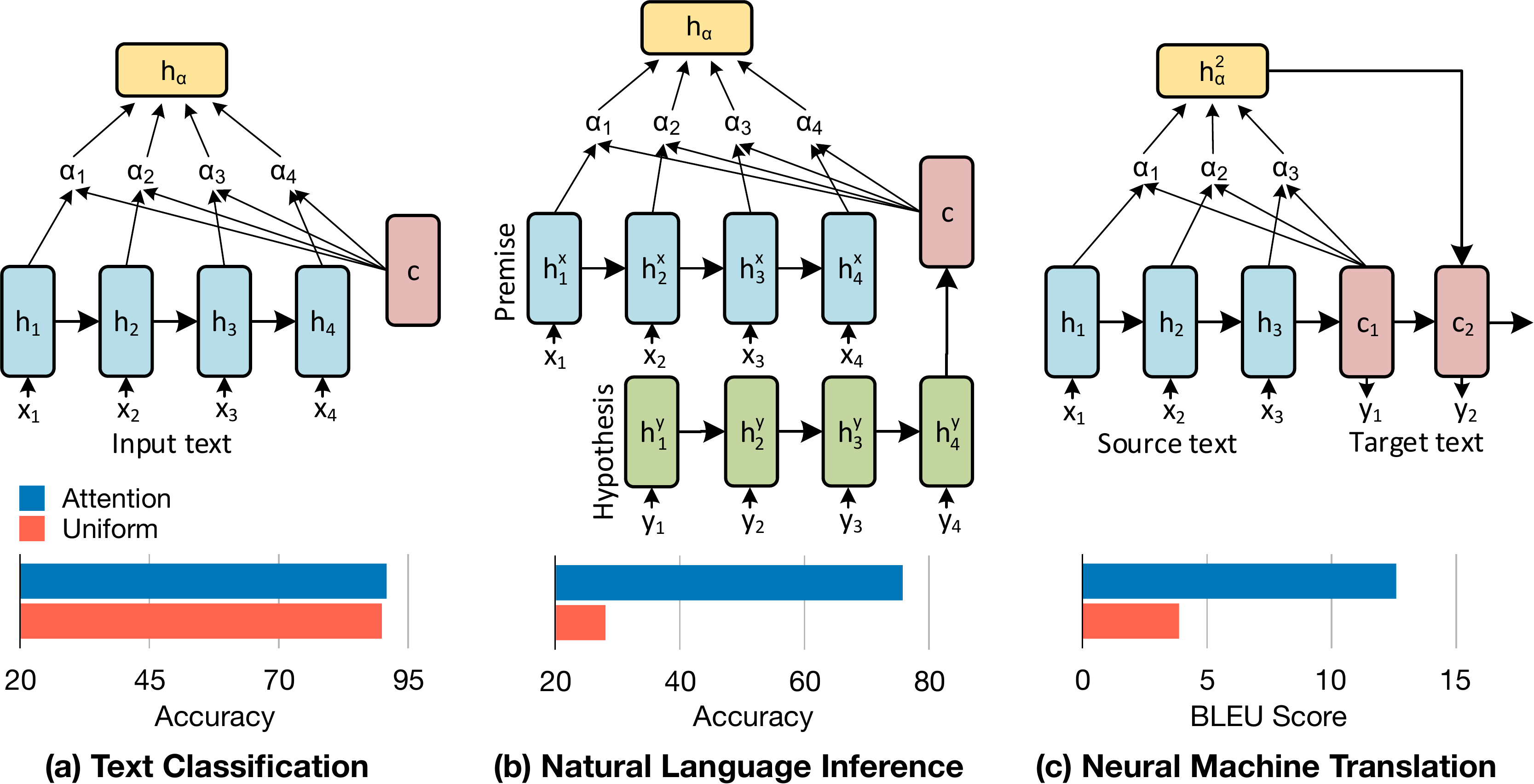}
	\caption{\label{fig:motivation_plot}Comparison of performance with and without neural attention on text classification (IMDB), Natural Language Inference tasks (SNLI) and Neural Machine Translation (News Commentary). Here, $\alpha$ and $c$ denote attention weights and context vector respectively. The results show that attention does not substantially effect performance on text classification. However, the same does not hold for other tasks. %Please refer to Section \ref{sec:introduction} for more details. %\mfar{Can we label these subfigures as (a), (b) and (c). So that they can referred to.} \sv{Done}
		%\sv{This figure can be removed.} \su{we can keep this figure, but we should add NMT also.}
	}
\end{figure*}

The conclusions of \cite{jain-wallace-2019-attention,smith_acl19} have been mostly 
%\mfar{why mostly? afaik it is only text classification} 
based on text classification experiments which might not generalize to several other NLP tasks. In Figure \ref{fig:motivation_plot}, we report the performance on text classification, Natural Language Inference (NLI) and Neural Machine Translation (NMT) of two models: one trained with neural attention and the other trained with attention weights fixed to a uniform distribution. The results show that the attention mechanism in text classification does not have an impact on the performance, thus, making inferences about interpretability of attention in these models might not be accurate. However, on tasks such as NLI and NMT
%freezing \mfar{replace "freezing" by uniform} 
uniform attention weights degrades the performance substantially, indicating that attention is a crucial component of the model for these tasks and hence the analysis of attention's interpretability here is more reasonable.

%In our work, we demonstrate several such fralities in the prior works and attempt to provide an explanation for their observations.

In comparison to the existing work on interpretability, we analyze attention mechanism on a more diverse set of NLP tasks that include text classification, pairwise text classification (such as NLI), and text generation tasks like neural machine translation (NMT). Moreover, we do not restrict ourselves to a single attention mechanism and also explore models with self-attention. %based models which have become predominant in NLP community. 
%\mfar{rewrite this line to exactly state HATT} 
For examining the interpretability of attention weights, we perform manual evaluation. %The key contributions of our paper can be summarized as follows: %\su{replace with something like "the key findings of our paper are" and rephrase the bullets below.}:
Our key contributions are:

%\mfar{Replace these to actually list the conclusions of the paper. These points currently are what you **did**, not what you **discovered**.}
\begin{enumerate}[itemsep=1pt,topsep=2pt,parsep=0pt,partopsep=0pt,leftmargin=20pt]
	\item We extend the analysis of attention mechanism in prior work to diverse NLP tasks and provide a comprehensive picture which alleviates seemingly contradicting observations.
	\item We identify the conditions when attention weights are interpretable and correlate with feature importance measures -- when they are computed using two vectors which are both functions of the input (Figure~\ref{fig:motivation_plot}b, c). We also explain why attention weights are not interpretable when the input has only single sequence (Figure~\ref{fig:motivation_plot}a), an observation made by \cite{jain-wallace-2019-attention}, by showing that they can be viewed as a gating unit.
% 	\item We show that attention weights are interpretable and correlate with feature importance measures when they are computed using two vectors which are both functions of the input (Figure~\ref{fig:motivation_plot}b, c) and not when one vector is a model parameter (Figure~\ref{fig:motivation_plot}a). %\sv{It looks confusing to read in the first go}%weights are not computed just based on the input and are crucial for model output. 
	%\item Through a battery of experiments with various attention mechanisms and self-attention based models we validate our hypothesis. We further reinforce our claims of attention interpretability through manual evaluation.
	\item We %analyze various attention mechanisms and self-attention based models and
	validate our hypothesis of interpretability of attention through manual evaluation.
\end{enumerate}

\section{Tasks and Datasets}
\label{sec:datasets}
We investigate the attention mechanism on the following three task categories.
\begin{enumerate}[itemsep=4pt,topsep=2pt,parsep=0pt,partopsep=0pt,leftmargin=20pt]
	\item \textbf{Single Sequence tasks} are those where the input consists of a single text sequence. For instance, in sentiment analysis, the task is to classify a review as positive or negative. This also includes other text classification tasks such as topic categorization.
	For the experiments, in this paper, we use three review rating datasets: (1) Stanford Sentiment Treebank \citep{sst_dataset}, (2) IMDB \citep{imdb_dataset} and (3) Yelp 2017\footnote{from www.yelp.com/dataset\_challenge} and one topic categorization dataset  AG News Corpus \textit{(business vs world)}.\footnote{www.di.unipi.it/~gulli/AG\_corpus\_of\_news\_articles.html}
	%Similar to \cite{jain-wallace-2019-attention}, we also experiment with two other single sequence tasks, i.e., 20 Newsgroup \textit{(hockey vs baseball)} \citep{20news_data} and datasets.

	\item \textbf{Pair Sequence tasks} comprise of a pair of text sequences as input. The tasks like NLI and question answering come under this category. NLI involves determining whether a \textit{hypothesis} entails, contradicts, or is undetermined given a \textit{premise}. We use Stanford Natural Language Inference (SNLI) \citep{snli_dataset} and Multi-Genre Natural Language Inference (MultiNLI) \citep{mnli_data} datasets for our analysis. For question answering, similar to \cite{jain-wallace-2019-attention}, we use CNN News Articles \citep{cnn_data} and  three tasks of the original babI dataset \citep{babi_dataset} in our experiments, i.e., using one, two and three supporting statements as the context for answering the questions.
	\item \textbf{Generation tasks} involve generating a sequence based on the input sequence. Neural Machine translation is an instance of generation task which comprises of translating a source text to a target language given translation pairs from a parallel corpus. For our experiments, we use three English-German datasets: Multi30k \citep{multi30k}, En-De News Commentary v11 from WMT16 translation task\footnote{http://www.statmt.org/wmt16/translation-task.html} and full En-De WMT13 dataset.
\end{enumerate}
%A summary statistics of all the datasets used in the paper is provided in Table \ref{tbl:datasets}.\mfar{Don't think table is needed.}

\section{Neural Attention Models}
\label{sec:models}
In this section, we give a brief overview of the neural attention-based models we analyze for different categories of tasks listed in Section \ref{sec:datasets}. The overall architecture for each category is shown in Fig \ref{fig:motivation_plot}.

\subsection{Single Sequence Models:}
\label{sec:single_models}
For single sequence tasks, we adopt the model architecture from \cite{jain-wallace-2019-attention,pinter_not_not_exp}. For a given input sequence $\bm{x} \in \real{T \times |V|}$, where $T$ and $|V|$ are the number of tokens and vocabulary size, we first represent each token with its $d$-dimensional GloVe embedding \cite{glove} to obtain $\bm{x}_e \in \real{T \times d}$. Next, we use a Bi-RNN encoder ($\bm{\mathrm{Enc}}$) to obtain an $m$-dimensional contextualized representation of tokens: $\bm{h} = \bm{\mathrm{Enc}}(\bm{x}_e) \in \real{T \times m}$. Then, we use the additive formulation of attention proposed by \cite{bahdanau_iclr14} for computing attention weights $\alpha_i$ for all tokens defined as:
\begin{equation}
\label{eqn:attention_comp}
\bm{u}_i = \mathrm{tanh}(\bm{W}\bm{h}_i + \bm{b}); \ \ \ \
\alpha_i = \dfrac{\mathrm{exp} (\bm{u}_i^{T} \bm{c})}{\sum_j{\mathrm{exp} (\bm{u}_j^{T} \bm{c})}},
\end{equation}
where $\bm{W} \in \real{d \times d'}, \bm{b}, \bm{c} \in \real{d'}$ are the parameters of the model.
%\mfar{Unclear what $\bm{c}$ is.} 
Finally, the weighted instance representation: $\bm{h}_{\alpha} = \sum_{i=1}^{T}\alpha_{i}\bm{h}_{i}$ is fed to a dense layer ($\bm{\mathrm{Dec}}$) followed by softmax to obtain prediction $\hat{y} = \sigma({\bm{\mathrm{Dec}}(\bm{h}_{\alpha})}) \in \real{|\mo{Y}|}$, where $|\mo{Y}|$ denotes the label set size.

We also analyze the hierarchical attention model \citep{hierarchical_atten}, which involves first computing attention over the tokens to obtain a sentence representation. This is followed by attention over sentences to obtain an instance representation $\bm{h}_{\alpha}$, which is fed to a dense layer for obtaining prediction $(\hat{y})$. At both word and sentence level the attention is computed similar to as defined in Equation \ref{eqn:attention_comp}.%\mfar{In single sequence tasks, there is only one sentence, so how is hierarchical attention used here?} \sv{In text classification, the input is a document which consists of multiple sentences. So first representation is learned for sentences then for the whole document. We can make it clear if it is not intuitive.}

\subsection{Pair Sequence Models:}
\label{sec:pair_models}
For pair sequence, the input consists of two text sequences: $\bm{x} \in \real{T_1 \times |V|}, \bm{y} \in \real{T_2 \times |V|}$ of length $T_1$ and $T_2$. In NLI, $\bm{x}$ indicates premise and $\bm{y}$ is hypothesis while in question answering, it is the question and paragraph respectively. Following \cite{snli_dataset}, we use two separate RNNs for encoding both the sequences to obtain $\{\bm{h}_1^{x}, ..., \bm{h}^{x}_{T_1}\}$ and $\{\bm{h}_1^{y}, ..., \bm{h}^{y}_{T_2}\}$. Now, similar to \cite{jain-wallace-2019-attention}, attention weight $\alpha_i$ over each token of $\bm{x}$ is computed as:
\begin{equation}
 \label{eqn:attenion_pair_1}
u_{i} = \mathrm{tanh}(\bm{W}_1\bm{h}^x_i +  \bm{W}_2\bm{h}^y_{T_2}); \ \ \ \
\alpha_i = \dfrac{\mathrm{exp} (\bm{u}_i^{T} \bm{c})}{\sum_j{\mathrm{exp} (\bm{u}_j^{T} \bm{c})}},
\end{equation}
where similar to Equation \ref{eqn:attention_comp}, $\bm{W}_1, \bm{W}_2 \in \real{d \times d'}$ denotes the projection matrices and $\bm{c} \in \real{d'}$ is a parameter vector. Finally, the representation obtained from a weighted sum of tokens in $\bm{x}$: $\bm{h}_\alpha = \sum_{i=1}^{T}\alpha_{i}\bm{h}_{i}^{x}$ is fed to a classifier for prediction.%\mfar{$\bm{c} = W_3 * h_{T2} + b_3?$} \sv{No, how did you get the equation?}
%\mfar{What is c here? unclear}.
%\su{we should have some good reason/rationale why we compare with \cite{rock_snli}}

We also explore a variant of the above attention proposed by \cite{rock_snli}. Instead of keeping the RNN encoders of both the sequences independent, \cite{rock_snli} use conditional encoding where the encoder of $\bm{y}$ is initialized with the final state of $\bm{x}$'s encoder. This allows the model to obtain a conditional encoding $\{\bm{h}'^{y}_1, ..., \bm{h}'^{y}_{T_2}\}$ of $\bm{y}$ given the sequence $\bm{x}$. Moreover, unlike the previous model, attention over the tokens of $\bm{x}$ is defined as follows:
\begin{equation}
\label{eqn:attenion_pair_2}
\bm{M} = \mathrm{tanh}(\bm{W}_1\bm{X} +  \bm{W}_2\bm{h}'^y_{T_2} \otimes \bm{e}_{T_1}); \ \ \ \
\bm{\alpha} = \mathrm{softmax}(\bm{w}^T\bm{M}),
\end{equation}
where $\bm{X} = [\bm{h}_1^{x}, ..., \bm{h}^{x}_{T_1}]$, $\bm{e}_{T_1} \in \real{T_1}$ is a vector of ones  and outer product $\bm{W}_2\bm{h}'^y_{T_2} \otimes \bm{e}_{T_1}$ denotes repeating linearly transformed $\bm{h}'^y_{T_2}$ as many times as words in the sequence $\bm{x}$ (i.e. $T_1$ times). %\mfar{Feels weird to represent hidden units of input x with $\bm{Y}$, either use $\bm{X}$ or something else?} \sv{Makes sense, corrected}

\subsection{Generation Task Models:}
\label{sec:generation_models}
In this paper, for generation tasks, we focus on Neural Machine Translation (NMT) problem which involves translating a given source text sentence $\bm{x} \in \real{T_1 \times |V_1|}$ to a sequence $\bm{y} \in \real{T_2 \times |V_2|}$ in the target language. 
%\mfar{V cannot be same for two languages.} 
The model comprises of two components: (a) an encoder which computes a representation for each source sentence and (b) a decoder which generates a target word at each time step. In this work, we utilize RNN based encoder and decoder models. For each input sentence $\bm{x}$, we first obtain a contextualized representation $\{\bm{h}_1, ..., \bm{h}_{T_1}\}$ of its tokens using a multi-layer Bi-RNN. Then, at each time step $t$, the decoder has a hidden state defined as
\[\bm{c}_t = f(\bm{c}_{t-1}, y_{t-1}, \bm{h}_{\alpha}^t),  \text{where} \ \  \bm{h}_{\alpha}^t = \sum_{i=1}^{T_1}\alpha_{t,i}\bm{h}_{i}.
\]
In our work, we compute $\alpha_{t,i}$ as proposed by \cite{bahdanau_iclr14} and \cite{luong_att}. The former computes attention weights using a feed-forward network, i.e., $\alpha_{t,i} = \bm{w}^T \mathrm{tanh}(\bm{W}[\bm{c}_t;h_i])$ while the latter define it simply as $\alpha_{t,i} = \bm{c}_t^T h_i$.
%\mfar{what is $g_i$}

\subsection{Self-Attention based models:}
\label{sec:self_attention_models}
We also examine \textit{self-attention} based models on all three categories of tasks.
% Given an embedded text sequence $\bm{X} \in \real{T \times d}$, self-attention involves computing attention between all pairs of tokens. Firstly, learned parameter matrices $\bm{W}_Q$, $\bm{W}_K$, $\bm{W}_V$ are used to compute query ($q_i = \bm{W}_Q x_i$), key ($k_i = \bm{W}_K x_i$) and value ($v_i = \bm{W}_V x_i$) vector respectively for each input token. The hidden representation for a token is given as: $h_i = \sum_{j=0}^{T}p(i \rightarrow j) v_j$, where $p(i \rightarrow j) = \mathrm{exp}(q_i^T k_j/\sqrt{d})$. In matrix form,
% \[
% \bm{H} = \mathrm{softmax}\left( \dfrac{\bm{Q}\bm{K}^T}{\sqrt{d}}\bm{V}\right), \text{where} \ \ \bm{Q} = \bm{W}_Q \bm{X}; \bm{K} = \bm{W}_K \bm{X}; \bm{V} = \bm{W}_V \bm{X}. 
% \]
For single and pair sequence tasks, we fine-tune pre-trained BERT \citep{bert_paper} model on the downstream task. In pair sequence tasks, instead of independently encoding each text, we concatenate both separated by a delimiter and pass it to BERT model. 
%\mfar{with no delimiter?} 
Finally, the embedding corresponding to \url{[CLS]} token is fed to a feed-forward network for prediction. For neural machine translation, we use Transformer model proposed by \cite{vaswani_attention_is_all} with \textit{base} configuration.

\section{Is Attention an Explanation?}
\label{sec:experiments}

%\mfar{We should stick to "interpretation" instead of "Explanation".}
In this section, we attempt to address the question: \textit{Is attention an explanation?} through a series of experiments which involve analyzing attention weights in a variety of models (\S\ref{sec:models}) on multiple tasks (\S\ref{sec:datasets}). Following \cite{jain-wallace-2019-attention}, we take the definition of \textit{explainability} of attention as: inputs with high attention weights are responsible for model output.
\cite{jain-wallace-2019-attention,smith_acl19} have extensively investigated this aspect for certain class of problems and have shown that attention does not provide an explanation. However, another series of work \citep{vig19,bert_attention_manning19,dipanjan_bert_acl} has shown that attention does encode several linguistic notions. In our work, we claim that the findings of both the line of work are consistent. We note that the observations of the former works can be explained based on the following proposition.

% \su{a lemma is something that is needed to prove a theorem later on. We do not have a theorem, so maybe we can call this ``Observation'' instead of Lemma.}
\begin{proposition}
\label{lemma:attention_is_gating}
Attention mechanism as defined in Equation \ref{eqn:attention_comp} as
\[
\bm{u}_i = \mathrm{tanh}(\bm{W}\bm{h}_i + \bm{b}); \ \ \ \
\alpha_i = \dfrac{\mathrm{exp} (\bm{u}_i^{T} \bm{c})}{\sum_j{\mathrm{exp} (\bm{u}_j^{T} \bm{c})}}
\]
for single sequence tasks can be interpreted as a gating unit in the network.
\end{proposition}
\textbf{Proof:} The attention weighted averaging computed in Equation \ref{eqn:attention_comp} for single sequence tasks can be interpreted as gating proposed by \cite{gated_cnn} which is defined as
\[
\bm{h}(\bm{X}) = f(\bm{X}) \odot \sigma(g(\bm{X})),
\]
where $\bm{X} \in \real{n \times d}$ is the input and $\odot$ denotes element-wise product between transformed input $f(\bm{X})$ and its computed gating scores $\sigma(g(\bm{X}))$. Equation \ref{eqn:attention_comp} can be reduced to the above form by taking $f$ as an identity function and defining $g(\bm{X}) = \bm{c}^T \mathrm{tanh}(\bm{W}\bm{X} + \bm{b})$ and replacing $\sigma$ with softmax. We note that the same reduction does not hold in the case of pair sequence and generation tasks where attention along with input also depends on another text sequence $\bm{Y}$ and current hidden state $\bm{c}_{t}$, respectively.
\qed

Based on the above proposition, we argue that weights learned in single sequence tasks cannot be interpreted as attention, and therefore, they do not reflect the reasoning behind the model's prediction. This justifies the observation that for the single sequence tasks examined in \citet{jain-wallace-2019-attention,smith_acl19}, attention weights do not correlate with feature importance measures and permuting them does not change the prediction of the model. In light of this observation, we revisit the explainability of attention weights by asking the following questions. %\sv{Commented the list of questions, it was looking repetitive}

% \begin{enumerate}
% 	\item[Q1.] How does altering attention weights affect model output on different tasks? (\S\ref{sec:results_alt_atten})
% 	\item[Q2.] Do attention weights correlate with feature importance measures? (\S\ref{sec:results_feature_imp})
% 	\item[Q3.] How permuting different layers of self-attention based models affect performance? (\S\ref{sec:results_self_attention})
% 	\item[Q4.] Are attention weights human interpretable? (\S\ref{sec:results_manual})
% \end{enumerate}

\subsection{How does altering attention weights affect model output on tasks?}
\label{sec:results_alt_atten}

In this section, we compare the performance of various attention mechanism described in \S\ref{sec:models} for different categories of tasks listed in \S\ref{sec:datasets}. For each model, we analyze its three variants defined as:
% \su{you should use maths from earlier sections here to explain what weights are randomly sampled, permuted etc.}
\begin{itemize}[itemsep=4pt,topsep=2pt,parsep=0pt,partopsep=0pt,leftmargin=20pt]
	\item \textbf{Uniform} denotes the case when all the inputs are given equal weights, i.e., $\alpha_i = 1/T, \ \forall i \in \{1, ..., T\}$. This is similar to the analysis performed by \cite{pinter_not_not_exp}. However, we consider two scenarios when the weights are kept fixed both during training and inference (Train+Infer) and only during inference (Infer).
	\item \textbf{Random} refers to the variant where all the weights are randomly sampled from a uniform distribution: $\alpha_i \sim U(0,1), \ \forall i \in \{1, ..., T\}$, this is followed by normalization. Similar to Uniform, we analyze both Train+Infer and Infer.
	\item \textbf{Permute} refers to the case when the learned attention weights are randomly permuted during inference, i.e., $\bm{\alpha} = \mathrm{shuffle}(\bm{\alpha})$. Unlike the previous two, here we restrict our analysis to only permuting during inference as Tensorflow currently does not support backpropagation with shuffle operation.\footnote{https://github.com/tensorflow/tensorflow/issues/6269}
\end{itemize}

\begin{table}[!t]
	\centering
	\small
	\begin{tabular}{lcccc}
		\toprule
		& \textbf{SST} & \textbf{IMDB} & \textbf{AG News} & \textbf{YELP} \\
		\midrule
		\cite{bahdanau_iclr14}  & $ 83.4 \pm 0.5  $ & $ 90.7 \pm 0.7   $ & $ 96.4 \pm 0.1  $ & $   66.7 \pm 0.1 $ \\
		\midrule
		Uniform ({\scriptsize Train+Infer / Infer})   & $ -1.0 \ $/$ -0.8 $ & $ -0.8 \ $/$ -6.3 $ & $ -0.1 \ $/$ -0.7 $ & $ -0.5 \ $/$ -6.3 $ \\
		Random  ({\scriptsize Train+Infer / Infer})   & $ -1.1 \ $/$ -0.9 $ & $ -0.6 \ $/$ -6.4 $ & $ -0.0  \ $/$ -0.7 $ & $ -0.4 \ $/$ -6.4 $ \\
		Permute ({\scriptsize Infer})         & $ -1.7 $ & $ -5.1 $ & $ -0.9 $ & $ -7.8 $ \\
		\midrule
		\midrule
		\cite{hierarchical_atten}  & $ 83.2 \pm 0.5   $ & $ 89.7 \pm 0.6   $ & $  96.1 \pm 0.2 $ & $    65.8 \pm 0.1 $ \\
		\midrule
		Uniform ({\scriptsize Train+Infer / Infer})   & $ -1.0 \ $/$ -0.8 $ & $ +0.2 \ $/$ -6.5 $ & $  +0.1 \ $/$  -1.5 $ & $ -0.7 \ $/$ -8.0 $ \\
		Random ({\scriptsize Train+Infer / Infer})    & $ -0.9 \ $/$ -1.0 $ & $ -1.2 \ $/$ -8.2 $ & $  -0.1 \ $/$  -1.8 $ & $ -3.0 \ $/$ -10.2 $ \\
		Permute ({\scriptsize Infer})         & $ -1.8 $ & $ -5.1 $ & $  -0.7 $ & $ -10.7 $ \\
		\bottomrule
	\end{tabular}
	\caption{\label{tbl:results_text_class}Evaluation results on single sequence tasks. We report the base performance of attention models and absolute change in accuracy for its variant. We note that across all datasets, degradation in performance on altering attention weights during inference is more compared to varying them during both training and inference. Overall, the change in performance is less compared to other tasks. Please refer to \S\ref{sec:results_alt_atten} for more details. %\mfar{Specify that Yang et al is HATT} \mfar{}specify that the rows here denote absolute loss in performance.
	} \vspace{-3mm}
\end{table}

%We report our results on single sequence, pair sequence and generation tasks in Table \re
\textbf{Effect on single sequence tasks:} The evaluation results on single sequence datasets: SST, IMDB, AG News, and YELP are presented in Table \ref{tbl:results_text_class}. We observe that Train+Infer case of Uniform and Random attentions gives around $0.5$ and $0.9$ average decrease in accuracy compared to the base model. However, in Infer scenario the degradation on average increases to $3.9$ and $4.5$ absolute points respectively. This is so because the model becomes more robust to handle altered weights in the former case. 
%when \su{``it encounters it'' reads wierd} during training. 
The reduction in performance from Permute comes around to $4.2$ across all datasets and models. The results support the observation of \cite{jain-wallace-2019-attention,smith_acl19} that alternating attention in text classification task does not have much effect on the model output. The slight decrease in performance can be attributed to corrupting the existing gating mechanism which has been shown to give some improvement \citep{gating_benefit,gated_cnn,srl_gcn}.
%\sv{Include that small reduction in performance because of removing gating mechanism}
% \mfar{Where are these numbers in the table?}

\textbf{Effect on pair sequence and generation tasks:} The results on pair sequence and generation tasks are summarized in Table \ref{tbl:results_nli_qa} and \ref{tbl:results_nmt}, respectively. Overall, we find that the degradation in performance from altering attention weights in case of pair sequence and generation tasks is much more substantial than single sequence tasks. For instance, in Uniform (Train+Infer), the average relative decrease in performance of single sequence tasks is $0.1 \%$ whereas in case of pair sequence and generation tasks it is $49.5 \%$ and $51.2 \%$ respectively. The results thereby validate our Proposition \ref{lemma:attention_is_gating} and show that altering attention does affect model output for a task where the attention layer cannot be modeled as a gating unit in the network.%attention weights are not a function of input itself. 
% denies the claim of prior works that altering attention has no effect . We note that the conclusion of prior works 

% \su{add a sentence statinghow this is adding to what we know from J and W etc.}

\begin{figure*}[h!]
	\centering
	\includegraphics[width=0.95\linewidth]{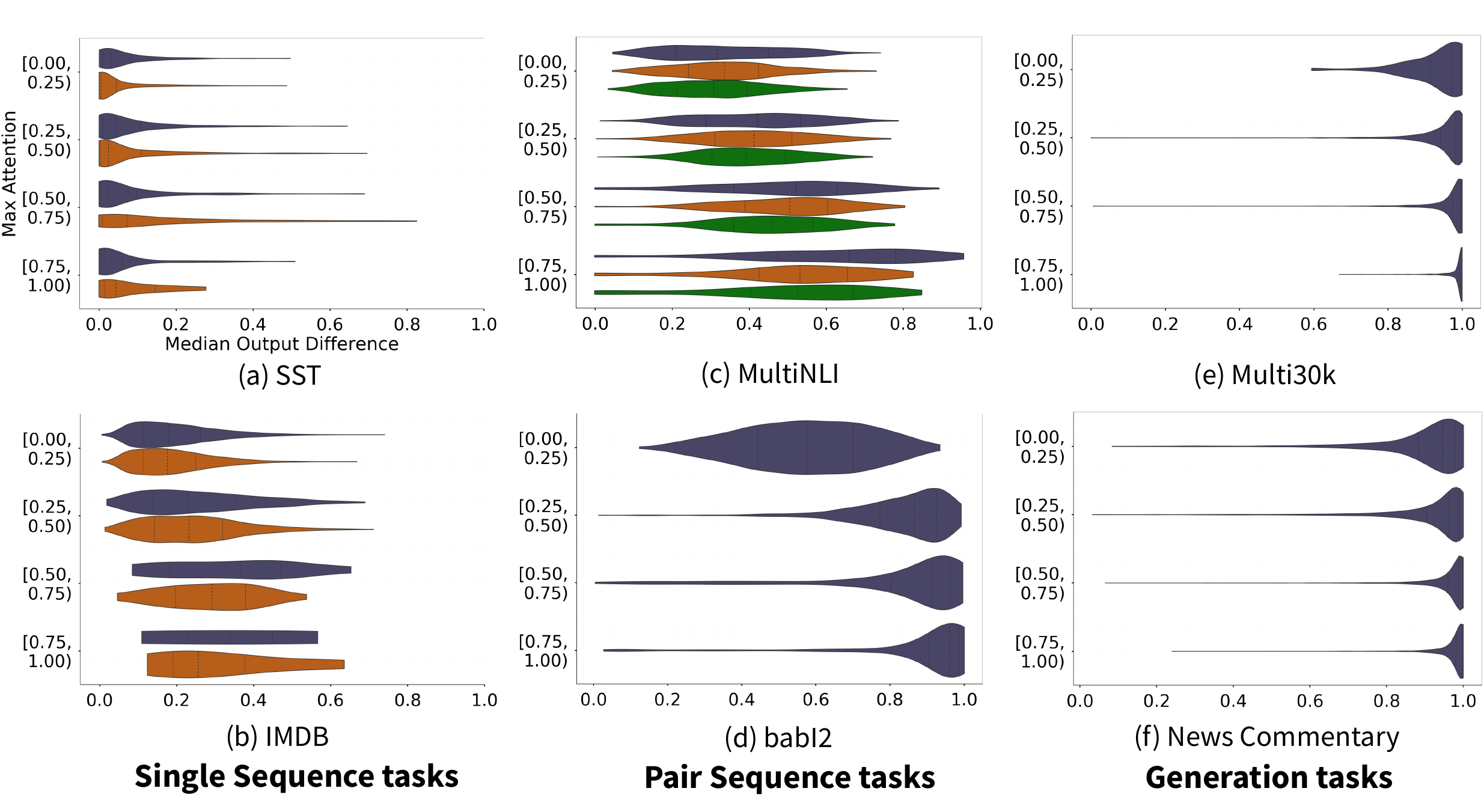}
	\caption{\label{fig:violin_plots}Relationship between maximum attention weight and median change in output on permuting attention weights. For single sequence tasks,  \textcolor[HTML]{7570b3}{$\blacksquare$}, \textcolor[HTML]{d95f02}{$\blacksquare$}   indicate negative and positive class. For MultiNLI, \textcolor[HTML]{7570b3}{$\blacksquare$}, \textcolor[HTML]{d95f02}{$\blacksquare$}, \textcolor[HTML]{1b9e77}{$\blacksquare$} denotes contradiction, entailment and neutral respectively. 
	The results reinforce the claim that altering attention weights in single sequence tasks does not have much effect on performance while the same does not hold with other tasks. Refer to \S\ref{sec:results_alt_atten} for details.}
\end{figure*}

\textbf{Visualizing the effect of permuting attention weights:} To further reinforce our claim, similar to \cite{jain-wallace-2019-attention}, we report the median of Total Variation Distance (TVD) between new and original prediction on permuting attention weights for each task. The TVD between two predictions $\hat{y}_1$ and $\hat{y}_2$ is defined as:
$
\text{TVD}(\hat{y}_1, \hat{y}_2) = \frac{1}{2} \sum_{i=1}^{|\mo{Y}|} |\hat{y}_{1i} - \hat{y}_{2i} |,
$
where $|\mo{Y}|$ denotes the total number of classes  in the problem. We use TVD for measuring the change in output distribution on permuting the attention weights. 
%\su{give short intuition behind what TVD is.}
In Figure \ref{fig:violin_plots}, we report the relationship between the maximum attention value and the median induced change in model output over $100$ permutations on all categories of tasks. For NMT task, we present change in output at the $25$th-percentile length of sentences for both datasets.  Overall, we find that for single sequence tasks even with the maximum attention weight in range $[0.75, 1.0]$, the change in prediction is considerably small (the violin plots are to the left of the figure) compared to the pair sequence and generation tasks (the violin plots are to the right of the figure).

\begin{table}[t]
	\centering
	\small
		\resizebox{\textwidth}{!}{
	\begin{tabular}{lcccccc}
		\toprule

& \textbf{SNLI} & \textbf{MultiNLI} & \textbf{CNN} & \textbf{babI 0} & \textbf{babI 1} & \textbf{babI 2}\\
\midrule
\cite{bahdanau_iclr14} & $75.7 \pm 0.3$ &    $61.1 \pm 0.1$ &    $63.4 \pm 0.8$ &    $96.1 \pm 4.3$ &    $95.8 \pm 0.3$ &    $92.8 \pm 0.1$ \\
\midrule
Uniform ({\scriptsize Train+Infer / Infer})                & $-41.8 \ $/$-42.9$ & $-26.6 \ $/$-28.7$ & $-30.8 \ $/$-55.9$ & $-44.4 \ $/$-63.4$ & $-47.4 \ $/$-60.4$ & $-48.4 \ $/$-62.1$ \\
Random ({\scriptsize Train+Infer / Infer})                  & $-41.6 \ $/$-43.1$ & $-26.7 \ $/$-28.6$ & $-30.9 \ $/$-55.9$ & $-45.0 \ $/$-62.0$ & $-47.3 \ $/$-60.4$ & $-49.9 \ $/$-62.2$ \\
Permute ({\scriptsize Infer}) & $-41.0$ & $-27.6$ & $-54.8$ & $-67.8$ & $-68.3$ & $-66.7$ \\
\midrule
\midrule
\cite{rock_snli}       & $78.1 \pm 0.2$ &    $62.4 \pm 0.6$ &    $63.6 \pm 0.6$ &    $98.6 \pm 1.6$ &    $96.2 \pm 0.9$ &    $93.2 \pm 0.1$ \\
\midrule
Uniform ({\scriptsize Train+Infer / Infer}) & $-44.2\ $/$-45.4$  & $-27.5\ $/$-30.3$ & $-30.8\ $/$-43.1$ & $-47.7\ $/$-67.8$ & $-47.9\ $/$-62.8$ & $-49.8\ $/$-60.9$ \\
Random ({\scriptsize Train+Infer / Infer}) & $-44.3\ $/$-44.9$  & $-27.9\ $/$-28.3$ & $-30.6\ $/$-43.3$ & $-47.5\ $/$-64.9$ & $-48.4\ $/$-63.3$ & $-49.8\ $/$-60.9$ \\
Permute ({\scriptsize Infer}) & $-41.7$ & $-29.2$ &  $-44.9$ &  $-68.8$ &  $-68.3$  & $-65.2$ \\
\bottomrule
\end{tabular}
}
\caption{\label{tbl:results_nli_qa}The performance comparison of attention based models and their variants on pair sequence tasks. We find that the degradation in performance is much more than single sequence tasks. }

\end{table}

\begin{table}[!t]
	\centering
	\small
	\begin{tabular}{lcc}
		\toprule

		Dataset     & \textbf{Multi30k} & \textbf{News Commentary} \\
\midrule
\cite{bahdanau_iclr14}             & $  31.3 \pm 0.1  $ & $ 12.6 \pm 0.1 $ \\
\midrule
Uniform ({\scriptsize Train+Infer / Infer}) & $ -10.4 \ $/$ -29.4 $ & $ -8.7 \ $/$ -11.8 $\\
Random ({\scriptsize Train+Infer / Infer})  & $ -10.1 \ $/$ -29.4 $ & $ -8.8 \ $/$ -11.9 $\\
Permute ({\scriptsize Infer})       & $ -29.7 $ & $ -12.1 $ \\
\midrule
\midrule
\cite{luong_att}                 & $  31.5 \pm 0.2  $ & $ 12.7 \pm 0.2 $ \\
\midrule
Uniform ({\scriptsize Train+Infer / Infer}) & $ -10.6 \ $/$ -29.7 $ & $ -8.8 \ $ /$ -12.0 $\\
Random ({\scriptsize Train+Infer / Infer})  & $ -10.3 \ $/$ -29.8 $ & $ -8.9 \ $ /$ -12.0 $\\
Permute ({\scriptsize Infer})       & $ -30.1 $ & $ -12.2 $ \\
		\bottomrule
	\end{tabular}
	\caption{\label{tbl:results_nmt}Evaluation results on neural machine translation. Similar to pair sequence tasks, we find that the deterioration in performance is much more substantial than single sequence tasks. Please refer to \S\ref{sec:results_alt_atten} for more details.} \vspace{-3mm}

\end{table}

\subsection{Do attention weights correlate with feature importance measures?}
\label{sec:results_feature_imp}
\begin{figure*}[t!]
	\centering
	\includegraphics[width=\linewidth]{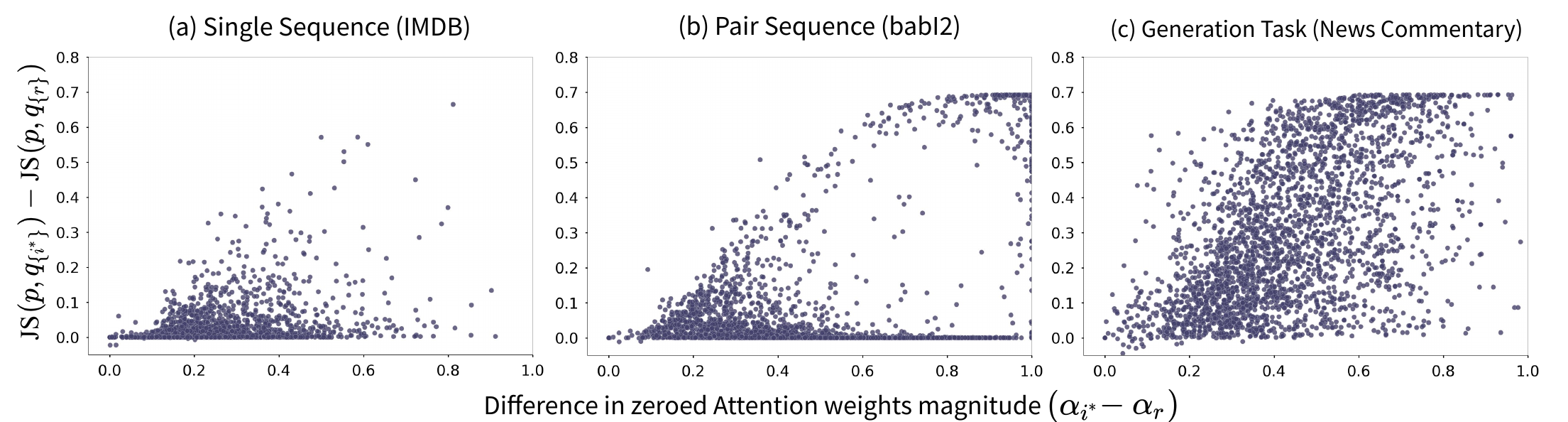}
	\caption{\label{fig:scatter_plot}Analysis of correlation between attention weights and feature importance measure. We report relationship between difference in zeroed attention weights and corresponding change in $JS$ divergence for different tasks. Please refer to \S\ref{sec:results_feature_imp} for more details.}
\end{figure*}

In this section, similar to the analysis of \cite{smith_acl19}, we investigate the importance of attention weights only when one weight is removed. Let $i^*$ be the input corresponding to the highest attention weights and let $r$ be any randomly selected input. We denote the original model's prediction as $\bm{p}$ and output after removing $i^*$ and $r$ input as $\bm{q}_{\{i^*\}}$ and $\bm{q}_{\{r\}}$ respectively. Now, to measure the impact of removing $i^*$ relative to any randomly chosen input $r$ on the model output, we compute the difference  of Jensen-Shannon ($\mathrm{JS}$) divergence between $\mathrm{JS}(\bm{p}, \bm{q}_{\{i^*\}})$ and $\mathrm{JS}(\bm{p}, \bm{q}_{\{r\}})$ given as:
$
\Delta \mathrm{JS} = \mathrm{JS}(\bm{p}, \bm{q}_{\{i^*\}}) - \mathrm{JS}(\bm{p}, \bm{q}_{\{r\}}).
$
The relationship between the difference of attention weights corresponding to $i^*$ and $r$, i.e., $\alpha_{i^*} - \alpha_{r}$ and $\Delta \mathrm{JS}$ for different tasks is presented in Figure \ref{fig:scatter_plot}. In general, we found that for single sequence tasks, the change $\mathrm{JS}$ divergence is small even for cases when the difference in attention weight is considerable. However, for pair sequence and generation tasks, there is a substantial change in the model output. 
%The results further support our thesis that attention in single sequence task does not play much role in its prediction.

 \vspace{-1mm}

\subsection{How permuting different layers of self-attention based models affect performance?}
\label{sec:results_self_attention}
\begin{figure*}[t!]
	\centering
	\includegraphics[width=0.9\linewidth]{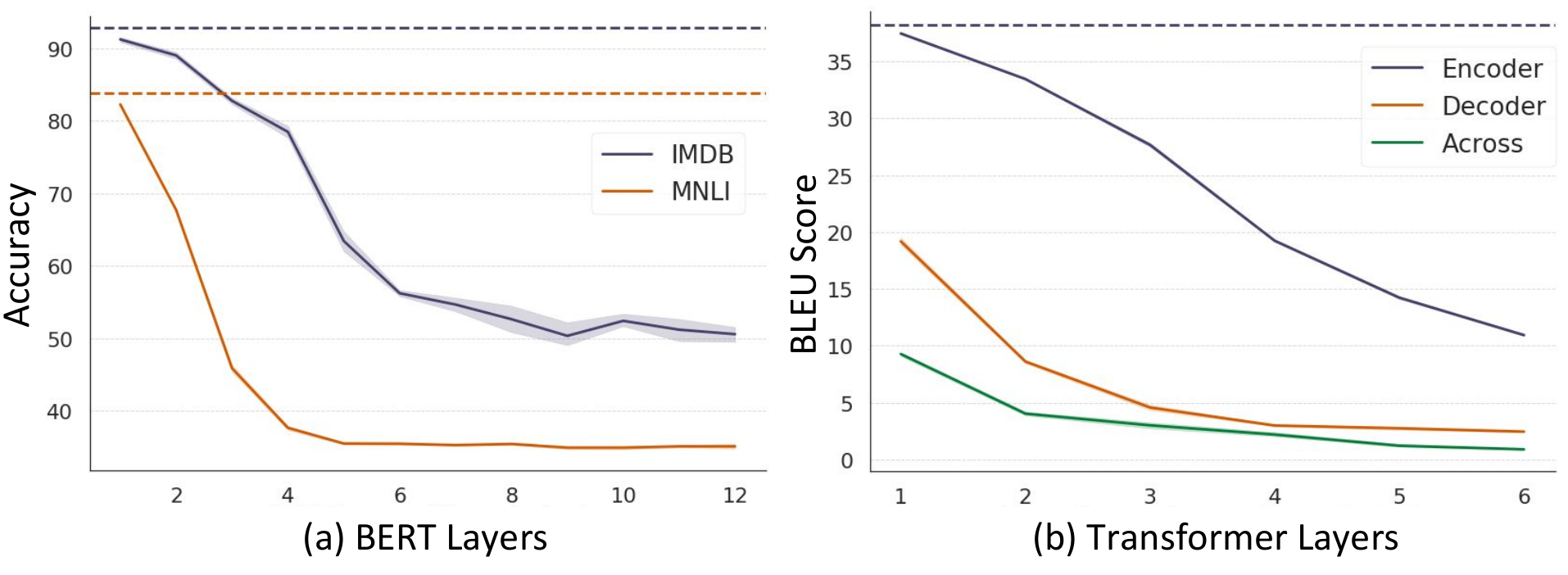}
	\caption{\label{fig:self_atten_plots}Performance comparison with permuting attention weights for different layers in self-attention based models. The results are reported on a representative instance of single sequence, pair sequence and generation tasks. The dotted lines denote the base performance on the task. Refer to \S\ref{sec:results_self_attention} for details. 
% 	\mfar{Lets' remove "cumulative" from the labels here and describe it only in text.}
	%Overall, we find that permuting weights in self-attention based models have a considerable effect on the performance even for the case of single sequence tasks which is unlike the case with standard neural attention.%\mfar{Can we make the same range for the BERT plots, say, 30-95} \mfar{Similarly for transformers, from 0-35 for both.} \mfar{We should include a line with the baseline (no change in attention) which will remain constant in both figures.}
	} \vspace{-2mm}
\end{figure*}

% \mfar{Explain what does ``cumulative'' mean. Its not described anywhere}
In this section, we analyze the importance of attention weights on the performance of self-attention based models as described in \S\ref{sec:self_attention_models}. We report the accuracy on single, and pair sequence tasks and BLEU score for NMT on WMT13 dataset on permuting the attention weights of layers cumulatively. For Transformer model, we analyze the effect of altering attention weights in encoder, decoder, and across encoder-decoder (denoted by Across). The results are presented in Figure \ref{fig:self_atten_plots}. Overall, we find that unlike the pattern observed in \S\ref{sec:results_alt_atten} and \S\ref{sec:results_feature_imp} for single sequence tasks, altering weights in self-attention based models does have a substantial effect on the performance. We note that this is because while computing attention weights over all tokens with respect to a given token, Proposition \ref{lemma:attention_is_gating} does not hold. 
% weights in self-attention are actually significant as they are required for computing representation for each token based on its neighbors. 
Thus, altering them does have an impact across all three tasks. We note that in the case of transformer model, altering the weights in the first step of Decoder and in Across has maximum effect as it almost stops the flow of information from encoder to decoder. %\mfar{It might be better to only show cumulative plots, lets see how this looks after the baseline line is included.} \sv{Yes, that looks like a good idea}

\begin{figure*}[t!]
	\centering
	\includegraphics[width=0.7\linewidth]{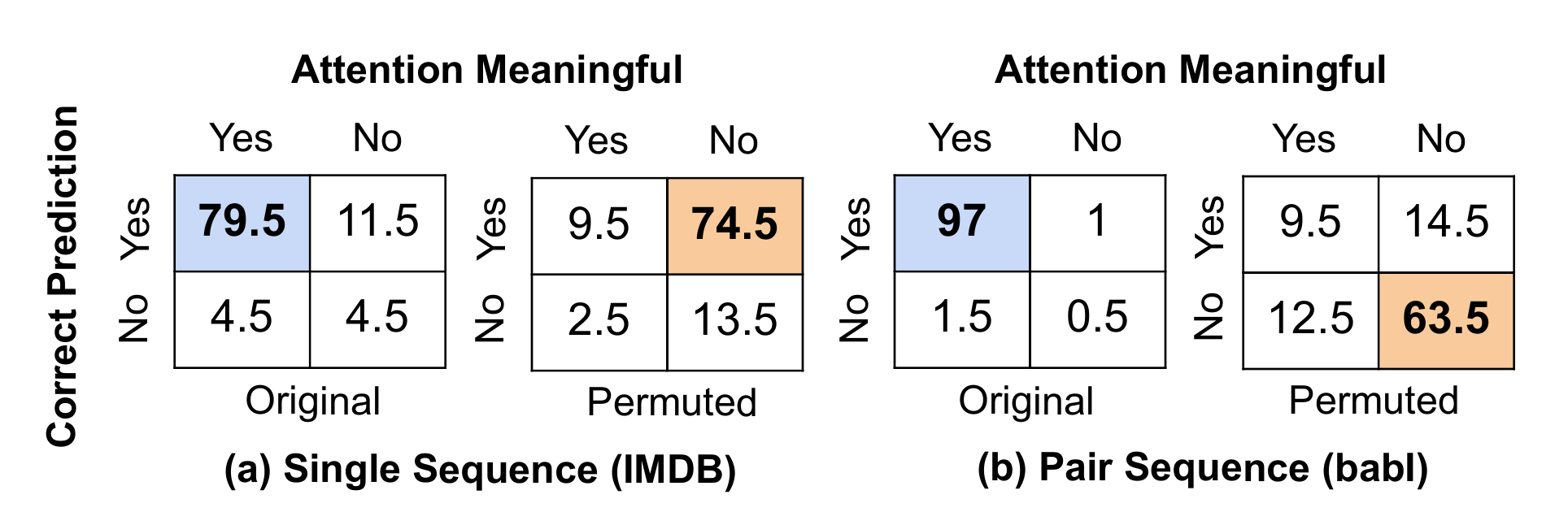}
	\caption{\label{fig:manual_eval}Manual evaluation of interpretability of attention weights on single and pair sequence tasks. Although with original weights the attention does remain interpretable on both tasks but in the case of single sequence tasks making it meaningless does not change the prediction substantially. However, the same does \textbf{not} hold with pair sequence tasks.} \vspace{-2mm} %Please refer to \S\ref{sec:results_manual} for more details.} 
\end{figure*} 

\vspace{-1mm}
\subsection{Are attention weights human interpretable?}
\vspace{-1mm}
\label{sec:results_manual}
% \mfar{Lets include 1-2 examples of what we have annotated as interpretable and what is not interpretable}

To determine if attention weights are human interpretable, here, we address the question of interpretability of attention weights by manually analyzing them on a representative dataset of single and pair sequence task. For each task, we randomly sample $100$ samples with original attention weights and $100$ with randomly permuted weights. Then, we shuffle all 200 samples together and present them to annotators for deciding whether the top three highest weighted words are relevant for the model's prediction. 

The overall results are reported in Figure \ref{fig:manual_eval}. Cohen's kappa score of inter-annotator agreement \citep{cohen1960} on IMDB and babI is $0.84$ and $0.82$, respectively, which shows near-perfect agreement \citep{Landis77}.
We find that in both single and pair sequence tasks, the attention weights in samples with original weights do make sense in general (highlighted with blue color). However, in the former case, the attention mechanism learns to give higher weights to tokens relevant to both kinds of sentiment. For instance, in ``\textit{This is a \underline{great} movie. \underline{Too} \underline{bad} it is not available on home video.}'', tokens \textit{great}, \textit{too}, and \textit{bad} get the highest weight. Such examples demonstrate that the attention mechanism in single sequence tasks works like a gating unit, as shown in \S\ref{sec:results_alt_atten}.

For permuted samples, in the case of single sequence, the prediction remains correct in majority although the attention weights were meaningless. For example, in \textit{``This movie was \underline{terrible} \textbf{.} the acting was \underline{lame} , \textbf{but} it 's hard to tell since \textbf{the} writing was so \underline{bad} .''}, the prediction remains the same on changing attention weights from \underline{underlined} to \textbf{bold} tokens. However, this does not hold with the pair sequence task. This shows that attention weights in single sequence tasks do not provide a reason for the prediction, which in the case of pairwise tasks, attention do reflect the reasoning behind model output.

\vspace{-1mm}
\section{Conclusion}
\vspace{-1mm}
In this paper, we addressed the seemingly contradictory viewpoint over explainability of attention weights in NLP. On the one hand, some works have demonstrated that attention weights are not interpretable, and altering them does not affect the model output while several others have shown that attention captures several linguistic notions in the model. We extend the analysis of prior works to diverse NLP tasks and demonstrate that attention weights are interpretable and are correlated with feature importance measures. However, this holds only for cases when attention weights are essential for model's prediction and cannot simply be reduced to a gating unit. Through a battery of experiments, we validate our claims and reinforce them through manual evaluation. 
\label{sec:conclusion}

\bibliography{references}
\bibliographystyle{iclr2020_conference}

% \appendix

\end{document}